# ENACTING SOCIAL ARGUMENTATIVE MACHINES IN SEMANTIC WIKIPEDIA

**Adrian Groza, Sergiu Indrie**
Technical University of Cluj-Napoca ,Romania
Adrian.Groza@cs.utcluj.ro

**ABSTRACT**

This research advocates the idea of combining argumentation theory with the social web technology, aiming to enact large scale or mass argumentation. The proposed framework allows mass-collaborative editing of structured arguments in the style of semantic wikipedia. The long term goal is to apply the abstract machinery of argumentation theory to more practical applications based on human generated arguments, such as deliberative democracy, business negotiation, or self-care. The ARGNET system was developed based on ther Semantic MediaWiki framework and on the Argument Interchange Format (AIF) ontology.

## 1 INTRODUCTION

The current research is the result of combining two visions: the Argumentative Web, proposed by Iyad Rahwan in 2007 [2], and the Social Machines envisaged by Tim Berners-Lee [14]. On the one hand, Argumentative Web is a large scale network of interconnected arguments created by human agents in a structured manner. The vision is to create an infrastructure for mass-collaborative editing of structured arguments. One desiderata is that Argumentative Web employs a unified extendable argumentation ontology. On the other hand, the Social Machines are the result of interaction of social web with the semantic technologies. The goal is to provide the users access to the next level of abstraction.

Building argumentation corpora is an active line of research [15], aiming to facilitate the use of argumentation theory in practical applications. The arguments are extracted either by i) automatic means, leading to the field of argumentation mining which processes free-text to detect natural occurring arguments, or ii) human based annotation, in which experts are employed to identify arguments and categorize them based on a set of argumentation schemes. At the moment few such argumentation corpora exist and they are limited in size. Two examples are Araucaria[1] with 641 arguments and ECHR (European Court of Human Rights) corpus with 257 arguments [9].

This research advocates the idea of integrating argumentation theory with the social web technology, aiming to enact large scale or mass argumentation by extending the conceptual work presented in [18] with the implemented system ARGNET [19]. Due to the large popularity of the well known Wikipedia, efforts were made to recreate that community spirit and propagate it to the Web 3.0. Semantic wikis are a subset of the participants of this movement, adding underneath the Wikipedia core a knowledge model capable of query and reasoning. In this paper, we implement a multi-user argumentation framework in Semantic MediaWiki based on the state of the art model, the Argument Interchange Format ontology [2].

## 2 TECHNICAL INSTRUMENTATION

### 2.1 Argumentation Schemes

For modelling the debate, we exploit the theoretical model of Walton based on argumentation schemes [1]. Argument schemes encapsulate common patterns of human reasoning such as: argument from position to know, argument from evidence, argument from sign, etc. Argumentation schemes are defined by the following items: a name, a set of premises (Ai), a conclusion (C) and a set of critical questions (CQi). Figure 1 details these attributes of Argument from expert opinion scheme. When a critical question is conveyed the conclusion is blocked until the issue risen by the CQ is clarified. CQs have the role to guide the argumentation process by providing the parties a subset from the most encountered possible counter-arguments.

---

[1] http://araucaria.computing.dundee.ac.uk/

```
┌─ Argument from expert opinion ≐ AS_EO ─
│ A₁ : E asserts that A is known to be true.
│ A₂ : E is an expert in domain D.
│ C : A may (plausibly) be taken to be true.
│ CQ₁ : Expertise Question − How credible is expert E as an expert source?
│ CQ₂ : Field Question − Is E an expert in the field that the assertion, A, is in?
│ CQ₃ : Opinion Question − Does E's testimony imply A?
│ CQ₄ : Trustworthiness Question − Is E reliable?
│ CQ₅ : Consistency Question − Is A consistent with the testimony of other experts?
│ CQ₆ : Backup Evidence Question − Is A supported by evidence?
```

**Figure 1**. Argument from expert opinion scheme.

### 2.2 Argument Ontology

To model the interaction between arguments, the AIF ontology is used. Defined originally by Chesnevar [3], the AIF represents a core ontology of argument-related concepts. There is a hierarchy of class nodes defined, where the two most important types are I-nodes which contain pieces of information, and S-nodes which represent a type of inference act. An information node I−node 2 NI represents passive information of an argument such as: claim, premise, data, locution, etc. A scheme node S − node 2 NS captures active information or domain-independent patterns of reasoning. The schemes are split in three disjoint sets, whose elements are: rule of inference schemes (RA−node), conflict application node (CA−node), preference application node (PA−node). RA−nodes are used to represent logical rules of inference such as modus ponens, defeasible modus ponens, modus tollens. CA−nodes represent declarative specifications of possible conflicts. PA−nodes allow to declaratively specify preferences among evaluated nodes.

An extension to the core AIF ontology is presented in [2] that would ease the representation of argument schemes. The schemes are provided with detailed descriptions (premise, conclusion and critical question descriptors) that must (or can in the case of critical questions) be fulfilled.

## 3 COMPUTATIONAL MODEL OF ARGUMENT

### 3.1 Argument Annotation

For the semantic annotation of arguments we use the semantic templates. of the Semantic Media Wiki (SMW) framework. SMW is an extension to the popular MediaWiki software package[2]. Templates are a simple way of reusing content or parameterizable structures. The advantage that SMW brings is that semantic annotations can be used inside templates, thus allowing consistent annotation without having to learn specific syntax.

In order to exploit the AIF ontology in the SMW, we used a mapping process [4] between the concepts and roles from the ontology, to the internal

---

[2] http://semantic-mediawiki.org/wiki/Semantic MediaWiki

structuring mechanisms available in the semantic media wiki. Thus, the class hierarchy, as well as class membership, are employed through the use of categories, object and datatype properties are defined through the use of the property namespace, whilst attribute values are inserted using semantic annotations. Adding a S-node into an existing hierarchy is done as follows

*[[imported from :: aif : S − node]][[category : Node]]*

which indicates that S−node is an element from the AIF vocabulary and that it is a subclass of Node. For creating a new conflict between arguments, in the Conflict Node category one has to enter:

*[[imported from: : aif: CA- node]][[category : SchemeNode]]*

After creating the two remaining top level subclasses of S-node, the scheme node hierarchy is complete.

In this wiki implementation, the pieces of information will be created as articles belonging to the I-node category. The I-node semantic template can be used with the syntax described in figure 2. Here, in order to create an argument the user can populate the following fields: summary, representing a short description of the argument, certainty representing the degree of belief the user would grant to the statement, text, theoretically unlimited text from a wiki article that encapsulates the content of the argument, supportURL, a link to the source of the information, and the context, represented as a list of terms from the imported ontologies (the cyc ontology and the foaf vocabulary are used in the example).

```
{{I-node
  |summary    = John said that it would rain tomorrow.
  |certainty  = High
  |text       = John, a weather man, carefully read ...
  |supportURL = http://www.theweathernet.com/weather
  |context    = [(1.0,cyc:weather),(0.8,foaf:topic)] }}
{{S-node
  | summary    = Considering the John's occupation and
                 the fact that he said it, proves it will rain.
  | certainty  = Very high
  | premises   = John's occupation, John said it rains
  | conclusion = On Friday (18.02.2010) it will rain.
  | supportURL = http://en.wikipedia.org/wiki/Inference
  | scheme     = Argument from position to know
  | topic      = Rain on Friday (18.02.2010)
  | default form = Argument from position to know}}
```

**Figure 2.** Semantic templates for I-node and S-node annotation.

S-nodes are a way of logically linking I-nodes. Regardless of the type of scheme used to build an argument, the structure can be summarized as a list of premises and a conclusion. A sample usage of the S-node template is displayed in figure 3. Here, the premises of the scheme are represented as a list of I nodes, the conclusion as a type of Node, scheme attribute encapsulates the type of the argumentation scheme used (i.e. argument from expert opinion). The default form is used internally for the graphical representation of the argument chains, whilst the topic field, representing the subject of the debate, is also defined as a list of terms provided by the imported ontologies.

The prototype system will aid the argument creation process by providing users with the option of selecting existing argument schemes. However, to encourage further project growth, users will have the ability to create new argument schemes. Figure 4 shows the usage of the template. The template necessary for scheme creation needs three attributes: a set of premise descriptions, a conclusion description and a set of critical questions.

Having this argument model, arguments can be annotated by selecting a descendant of S-node using an appropriate template, selecting the existing Node types for premise and conclusion, and choosing a descendant specific argument scheme. The existing arguments are linked in argument networks based on the following actions: i) creation, ii) infer, using RA node, iii) support, using PA-node, and iv) attack, based on CA-node.

### 3.2. Argument Reasoning

The Jena tool is exploited to perform the following reasoning tasks: 1) computing the argument validity: the state of the argument is computed from its credibility value; 2) generate explanations: chaining of argument content; 3) computing the degree of contradiction: measure of argument sub-network inconsistency or disapproval. An ontology model is created from the RDF file which serves as data for building an argument tree having the query element as root. This tree will contain all interaction with the query argument: argument components and all arguments inferring, supporting or attacking the argument or its descendants in the tree.

The state of the argument is computed from its credibility value. In order to establish concepts like argument validity and explanation we need to define metrics for evaluating the degree of argument support.

*User-defined certainty* ($\gamma$) consists of a numeric value selected by the user, symbolizing the certainty attached to the current Node. Here, the certainty values ± defined by the user are 1 for very low, 2 for low, 5 for average, 7 for high, and 9 for very high. To be able to determine argument validity we must define a function for calculating credibility: credibility(node) = $\gamma c+\upsilon u+\mu m+\alpha a+\sigma p+\sigma s$ where $\gamma$, $\upsilon$, $\mu$, $\alpha$, $\sigma$ and $\sigma$ are the weighted factors and c, u, m, a, p and s are the certainty, usage (the number of participations in other arguments), minimum support (the minimum credibility value of a supporting S-node or premise), conflict attacks (a count of CA-nodes with the current node as target), preference supports (a count of PA-nodes with the current node as target) and scheme type values (the importance given to each argument scheme) of the targeted node.

*Node usage* ($\upsilon$). An important factor in large-scale multi-user argumentation systems is element reuse. The idea is the more an argument is used, the more trustworthy it becomes. In order to capture this line of thought we keep track of element participation in arguments. In the case of I-nodes, we count the number of arguments in which it acts as a premise or as a conclusion. From this value we subtract the ones where the I-node is a conclusion for a preference or conflict node as those are not situations that support the item's credibility. Unlike I-nodes, S-nodes cannot take the role of premises, thus in this case we count the number of arguments in which the S-node is a conclusion, subtracting the ones where it is a conclusion for a PA-node or CA node.

*PA-nodes* ($\sigma$) are a means of allowing users to give extra support to an argumentation element. Using this concept, we can increase the credibility of facts (I-nodes) or reasoning acts (S-nodes). Thus, the PA-nodes factor is computed by summing up the credibility values of the number of preference nodes that have the current node as a conclusion. The final system allows parameter selection as well as weight attribution when computing the credibility computation.

*CA-nodes* ($\alpha$) present the user with the capability of contradicting arguments or statements. Like its opposite, the preference node, the CA node can negatively influence the credibility of nodes. This value is obtained by counting the number of conflict nodes that have the current node as conclusion.

*Minimum support* ($\mu$) contributes to the propagation of the credibility along the chain of arguments. The idea behind this factor is according to the weakest link principle, stating that an element is only as good as its lowest values support. For example, in the case of S-nodes, the node's credibility will take into consideration the smallest credibility of its premises, while in the case of a conclusion its credibility is computed using the minimum credibility of its supporting arguments of type RA-node. When calculating the minimum support, PA nodes and CA-nodes are not considered because their influence takes place at their specific factors.

*Scheme* ($\sigma$): Certain schemes, such as Argument from expert opinion might be valued highly than Argument from example, due to the reliability of the intrinsic source of the argument. Considering that assigning values to these schemes could be domain dependant or even improperly used, special configuration pages will be available.

By creating an extensible Java model that lets developers write simple classes that modify the credibility function and allowing users to specify the weight of each factor, we maintain a flexible method of evaluating arguments. Argument validity is defined as good credibility, thus determining validity can be summed up calculating a node's credibility and deciding whether it is above or below a balance point. In case the credibility is an integer, the balance point can be zero and the validity defined as a positive or negative number.

Explanation generation for accepting or defeating an argument rests on the computation of element credibility. Since several approaches exist for providing explanations, in this implementation we opt for the *best explanation* strategy. Thus, if one wishes to obtain the explanation of an argument, the system must evaluate all nodes in the argument tree with the requested argument as root and select the most credible path from the root to a descendant. A more technical description would be that the algorithm starts from the root of the tree (the requested claim) and it selects recursively the most credible child until there are no more children. Once it has found a path, it concatenates the summaries of all passing nodes in a reverse order.

The degree of contradiction *dc* in a topic or in the argument network can be stated in two ways. A simpler choice would be $dc(c, r, p) = c/(r + p)$, where $c$ is the number of conflict nodes, $r$ is the number of rule nodes and $p$ is the number of preference nodes. This method does not take into consideration the credibility values of each argument, therefore a more precise equation would be:

$$dc(c, r, p) = \frac{\sum_{i=1}^{n_c} credibility(c_i)}{\sum_{i=1}^{n_r} credibility(r_i) + \sum_{i=1}^{n_p} credibility(p_i)}$$

where $c$, $r$ and $p$ are vectors of conflict, rule and preference nodes, and $n_c$, $n_r$ and $n_p$ are the vector sizes.

### 3.3 Querying the Argument Corpus

We extend the querying capabilities provided by the Semantic MediaWiki with specific argument related capabilities. To identify the most adequate argumentation chain, both sources of annotations are exploited: i) argument annotations based on the AIF ontology and its argumentation schemes and ii) term annotations based on the ontologies and vocabularies imported into the Semantic Media Wiki framework. The following specific queries are considered:

*Search by AIF nodes*: The user can search the arguments created based on specific nodes in the AIF ontology. For instance, given the Implication node as a subclass of the Scheme application node:

[[imported from :: aif : Implication]][[category:S−node]]

one query might be: <Give all the implications of accepting argument X>.

*Search by scheme*: For this case, only specific patterns of arguments are accepted to contribute to the conclusion: <Give only the arguments from expert opinion for supporting the argument "undercooked food is not recommended for pregnant women"> or <List all the consequences based on practical reasoning scheme for the argument "use Linux on Servers">. In the first query the argument from expert opinion was used as filter, whilst for the second query, the argument from practical reasoning was employed.

*Search by wikipedia metadata*: Specific wiki related terms can be used to limit or refine the searching domain, such as i) user: <Give all the arguments from the user Y against argument X >, ii) data: <Give all the arguments posted from yesterday against "pollution">, or iii) location: <Give all the arguments of the users from Haiti against "ONU support">.

*Search by domain*: This search implies intensive reasoning on domain ontologies. Each semantic wiki page can be annotated with the domain to which belongs. Suppose that an argument against using doping substances in football exists. Based on the terminological box Football v Sport the system is able to include this argument in the answers list to the more general query <List all the arguments against the argument "doping in sport">.

*Search by degree of support*: Each dispute is characterized by a specific standard of proof, representing the willingness of the parties to accept unreliable arguments. Only the arguments whose degree of support satisfies the minimum threshold of the current dispute will be considered.

*Filter by context*: The acceptance of an argument is a combination of intrinsic and extrinsic factors. A successful argument in a context might have no relevance in another. In order to effectively support a consequent, flexible control must be exercised over the extrinsic factors of argumentation by providing a context [8]. The context is encapsulated as a list of terms from the ontologies imported in the Semantic Media Wiki. Thus, reasoning on context is similar to reasoning on the domain ontologies. One advantage is that the context helps agents to discover the available means of persuasion for the current debate.

The context of an argument can be seen as representing subjective perspectives on the argument. Several contextual dimensions can be formalized for a general argument such as dialectical context, intentional context or social context. By making use of wikipedia technology, the social context is of particular interest here. It encapsulates the human factors related to the context, which might refer to: information on the user (knowledge of habits, emotional state), social environment (co-location of friends, social interaction), cultural issues (e.g. acquisition of context), relationship between the specific arguments and the plans of the arguers.

### 3.4. Graphical Representation

This internal structure allows the construction of a graphic module that will use the Graph MediaWiki extension[3]. This visual plugin defines a complex graph description language through which it can render graphical representations in different formats such as ASCII, HTML, SVG or Graphviz compatible formats. An example of how a simple argument would be described using this extension is shown is figure 3, where two nodes are created for representing the *premise1* and *premise2*, linked with support-like arrows (--->) with the consequent *conclusion*, based on an S-node scheme.

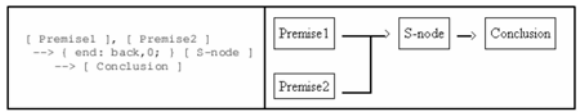

**Figure 3.** Graphical description language of a simple argument.

## 4  The ARGNET SYSTEM

The system architecture depicted in figure 4 can be viewed as two sections merging to create a distributed semantic argumentation system. On one hand, from the technical viewpoint, the design relies on the core MediaWiki installation which takes care of the distributed aspect of the application as well as the presentation layer by providing a web interface with a database backup. This core is converted into a semantic system using the SemanticMediaWiki extension which allows annotation and thus further reuse of arguments exported in the RDF syntax. To increase usability, the Semantic Forms extension was added on top, aiming to ease semantic template usage with the help of HTML forms. The Graph extension facilitates the presentation of argumentation chains.

On the other hand, from the conceptual point of view, Argnet has its foundation in Walton's argumentation schemes and their integration in the AIF. Using SMW we managed to model the AIF concepts as semantic templates. At this point the two branches merge with the help of our MediaWiki extension, customizable argumentation functions, and the Jena framework used for argument reasoning. This extension takes the user question as input, exports the wiki knowledge in RDF format and uses the Java reasoning application to supply the answer. We used Jena to read the wiki export and obtain information about element membership to create an internal model of the argument tree and provide answers such as argument validity, explanation, contradiction degree, and the graph description.

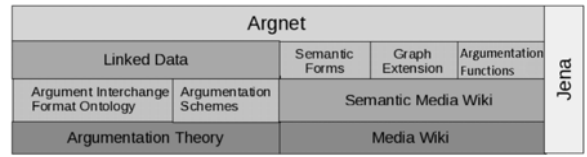

**Figure 4.** System architecture.

## 5  RUNNING SCENARIO

This section presents a simple usage scenario and shows the application results. The discussion is based on the network of arguments displayed in figure 5. Here, the information nodes are white, rule nodes are green, conflict nodes depicted with red, and preference nodes with blue.

### 4.1 Network construction

Suppose that Jim wishes to create an argument about the fact that good software costs more. He decides on what argument scheme fits best and chooses Argument from position to know. He builds the appropriate premises, conclusion and rule node (argument 1, where the credibility high is attached to the I-node <*Good software costs more*>). Another argument is provided by Sally based on the Argument from sign scheme, stating that <*Protege is developed in Java*> and considering that <*Java applications are usually free*>, so is Protege (argument 2). Consider the Sally sees on the wikipedia the claim that <*Good software costs*> and the fact that <*Protege is a good and free piece of software*>. Considering the fact that the claim is also true she creates a CA-node, attacking the RA-node of the argument 1 instead of its conclusion. Steve thinks <*Protege is typical of good Java software*> and builds argument 4 by re-using the conclusion of the argument 2. Lastly, Tom, a Java passionate, adds a PA-node (argument 5) on the argument that <*Java applications are good*>, using one of the premises of the argument 2.

### 4.2 Network processing

Before one begins any processing, he or she must establish certain parameters needed for calculating node credibility. The weights of the argument schemes used in this scenario are: 2 for *Argument from example*, 3 for *Argument from sign*, 4 for *Argument from position to know*, 3 for *Preference* and 3 for *Conflict*. The weights of the credibility factors: 0.02 for *user defined certainty* ($\gamma$), 0.7 for *node usage* ($\upsilon$), 1.5 for *PA-nodes* ($\sigma$), -1.5 for *CA nodes* ($\alpha$), 0.18 for *minimum support* ($\mu$) and 0.1 for *argument scheme* ($\sigma$).

---

[3] http://bloodgate.com/perl/graph/

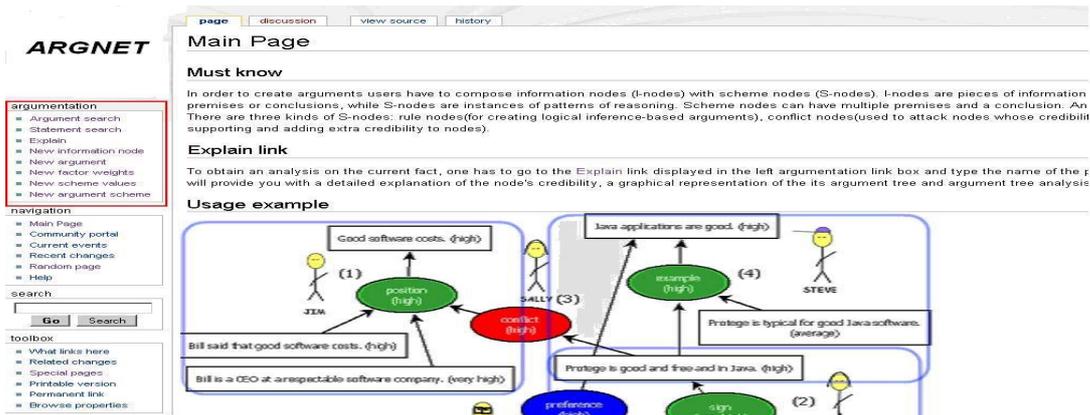

**Figure 5.** Enacting social argumentative machines.

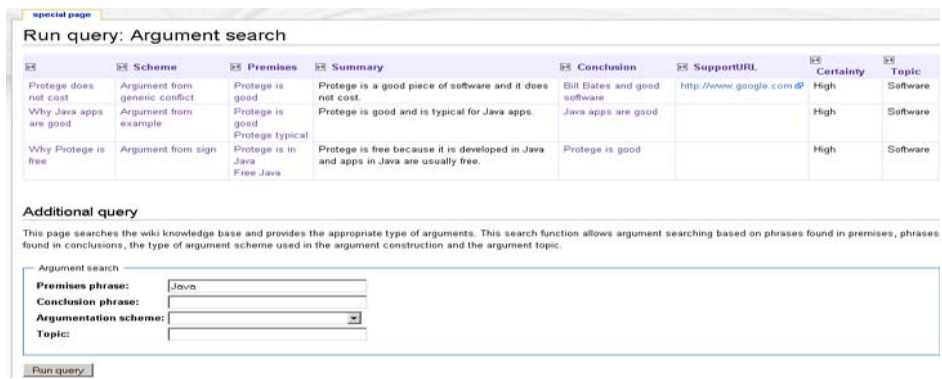

**Figure 7** Querying the argument corpus.

4.2.1 Argument validity:

If one would ask the validity of the argument 1 from figure 5 (using the interface illustrated in figure 6), the system calculates all the credibility values from the claim to its descendants. After evaluating all credibility values, the claim's validity fails with a value of -1.08. The conclusion fails mainly because of the strong conflict node (credibility = 1.59), supported by reusable information (reusability being a heavily weighed factor).

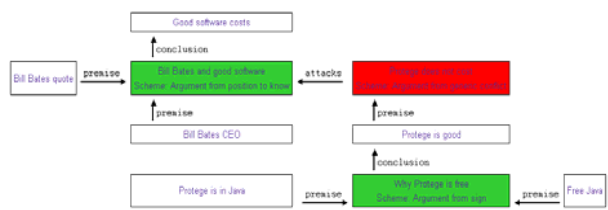

**Figure 6.** Visualisation of the argument chain.

4.2.2 Explanation

The generation of explanation also depends on credibility values. In order to build an explanation, the system connects argument components found on the most credible chain of arguments from the claim to one of its descendants. In this case the explanation for the argument <Good software costs> would be constructed from the premise of the argument 3 and from the antecedents of the argument 2. The ARGNET system automatically generates the graphical representation of the argument chain (figure 7) and the corresponding explanation for the conclusion validity (figure 8).

| Argument | Good software costs. |
|---|---|
| Status | The fact that Good software costs is not sufficiently supported. |
| Explanation | The facts that Java apps are usually free and that Protege is developed in Java indicate that Protege is good and free and in Java, which is in conflict with the current argumentation claim. |

**Figure 8.** Generating argument explanation.

4.2.3 Contradiction Degree:

This degree will be computed with the help of equation (1) and equation 1. Using the first one we count one conflict node, three rule nodes and one preference node, thus $cd(c, r, p) = 0.25$. By making use of the second equation we obtain the result $cd(c, r, p) = 0.53$.

# 6 ESTIMATED IMPACT

The benefits of enacting large scale argumentation are related to: i) augmenting human collaboration and argumentation by appropriate technologies; ii) extending the argumentative web towards a Pragmatic Web infrastructure for collaborative human-computer argument networks; iii) enhancing an individual's reasoning capabilities by increasing visibility, handle information overload, and providing users with re-usable patterns of argumentation. The current trend of forums, blogging, on-line debates is a positive social factor in the spirit of the current research [17]. The technology is seen as an important part in the argumentation process [7], with an exciting impact on several domains such as:

*Deliberative democracy* (e-Government, e-Administration). It involves dialog with the public and it requires many feedbacks, which must present themselves in a structured manner in order to be effectively processed and taken into consideration by the decision factors. The opposite direction, in which the local leaders justify their decision, is important for increasing transparency in e-democracy. The system helps when building multiple views of problems and resources among the following key actors: government and institutions, planners and technical experts, community. Services supporting structured argumentation impact e-government in: i) increasing transparency by providing structured and more clear justifications of decisions; ii) collecting relevant and motivated ideas from citizens in the form of structured public opinions; iii) supporting multiple views representation on an issue and public debates before norms adaptation. We anticipate the emergence of clusters of structured debates, in the context in which the technology for deploying structured government data in wikipedia is an intense research area[4].

*E-commerce*. The consumers can obtain more accurate information related to specific items. For instance, when one wants to buy a specific car, arguments pro and against can be browsed. In this line, our approach is complementary to field of "opinion mining", with the supplementary advantage of fetching the framework with more structured data.

*Self-care*. A lot of research centered on applications of argumentation in medicine (such as risk assessment or treatment planning) has led to a comprehensive view of argumentation as a form of evidential reasoning. By accessing arguments provided by related patients, one benefit is that the patient is helped to understand his or her health state.

# 7 DISCUSSIONS AND RELATED WORK

In our view, argumentation frameworks should have the ability to integrate the domain oriented aspect with the capability of re using arguments in different contexts. The domain oriented characteristic of the argumentative debates is supported by the following reasons: i) usually the participants of a debate share common interests; ii) standards of evaluating arguments are also domain dependent [11]. One benefit of re-using existing arguments in different contexts is that it provides a bottom up approach for developing large networks of interconnected arguments. We advocate the idea of combining argumentation theory with the social web technology aiming to enact large scale or mass argumentation. At the moment, there exists a primordial soup of prototype systems that support argumentation: The current trend consists in developing hybrid approaches that combine the advantages of formal (logic-based) and informal (argumentation scheme based, diagramming reasoning) ideas [5]. Among the variety of prototype systems that support argumentation: Rationale [12], Araucaria [10], Carneades [6], Reasonable, Magtalo (Multi-Agent Argumentation, Logic and Opinion), Aver (Argument visualization for evidential reasoning), Compendium[5], none seem to overcome a minimum number of users.

Debatepedia [6] is a wiki encyclopedia of arguments and debate related materials, including domains such as critical thinking, education, deliberative democracy. It provides a searchable repository of debates and the corresponding arguments supporting them, but without any formalization. We address the issue of large scale argumentation from a more structured viewpoint, exploiting the benefits of semantic technologies for enhancing query capabilities of the system.

Araucaria analyzes arguments based on diagrammatic reasoning, which also deploys a repository of debates. It provides a user-customizable set of schemes with which the human agent can analyze arguments and save them in the Argument Markup Language format. One output of the proposed framework is a large scale argumentation corpus semantically annotated. One difference is that both Araucaria and ECHR corpus are annotated by experts, which is not necessarily the case in our framework. The existence of such large size argument base will trigger the use of machine learning techniques in the argumentation field. We make use of the AIF ontology, which represents the state of the art standard at the moment when formalizing arguments and we exploits the Jena reasoning capabilities to provide more accurate

---

[4] Specific use cases are detailed at http://www.w3.org/egov/wiki/Use Cases

[5] http://compendium.open.ac.uk
[6] http://wiki.idebate.org, launched on October 2007

answers when searching within the argument base.

Inference engines for computing the acceptability of arguments have been developed under the ASPIC project. To prove the AIF concepts in this prototype, each node has a degree of support (dos 2 [0, 1]) attribute. Also, the computation of the dos of a conclusion based on the dos of its premises is based on the weakest link principle, according to which the dos of the consequent equals the minimum degree of support of its antecedents. In the large scale, open context of WWAW, these attributes might not suffice due to: i) standards of evaluating arguments are domain dependent; ii) the applicable principle of inference for computing the reliance on an argument may change during the course of argumentation. iii) the applicable principle of inference depends on the current context; iv) different principles require different attributes attached to the premises instead of the degree of support, such as fuzzy numbers or rough intervals.

**Table 1.** Mass Argumentation Tools

| FEATURE | ARGNET | DEBATEPEDIA | ARGUMENTUM | DEBATEGRAPH |
|---|---|---|---|---|
| Argument model | ASs and AIF ontology | PRO/CON structure | Support/Oppose idea | Natural language debate |
| Semantic annotation | Yes | No | No | No |
| Graphical representation | Wiki pages and graph-based representation of he chains | Wiki pages with a PRO/CON structure | List of text articles | Graphic environment using concept maps |
| Degree of support | User defined | None | None | Element rating |
| Reasoning support | Validity, explanation and contradiction degree | None | None | None |
| Query capabilities | Use of SMW query language | Relying only on the MediaWiki mechanism | Implementation specific (by text, position, date) | Simple text based |
| User contribution | Only created arguments list (MediaWiki) | Simple user pages | Evaluation of user's contribution and comparison | Authors of elements displayed |
| Connectivity | RDF export | Simple MediaWiki export | None | RSS feeds |

Therefore, we designed a flexible framework in which the users can choose from a set of possible argument evaluation strategies, aiming to map the standard of proof with the current domain of the dispute.

In order to provide a higher view of our system's characteristics we compare it with three state of the art large scale mass argumentation systems: Debatepedia, Argumentum and Debategraph. The following features were considered (see table I): i) the argument model; ii) whether or not the system uses semantic annotation; iii) the graphical representation; iv) the possibility of assigning a degree of support to an element; v) the reasoning support provided by the software, this extending the usability of the system from an argumentation support system to an argumentation-capable system; vi) the query capabilities that help users to efficiently locate information; vii) the role and value of users, which can constitute into a credibility factor itself; viii) and the connectivity, an essential factor in the context of mass argumentation.

## 8 CONCLUSION

The proposed ARGNET framework allows mass-collaborative editing of structured arguments in the style of semantic wikipedia. In this study we have shown that using a prolific environment such as SMW as an argumentation platform could have numerous benefits and present a stable, yet flexible foundation for further development. The AIF ontology provides the opportunity of integrating software agents [16] within argumentative web, where the agents use the argument reasoning capabilities of our framework and structured facts to build smart argument spaces [20] in Semantic Wikipedia.

## ACKNOWLEDGEMENT

This work was supported by the grant ID_170/672 from the National Research Council of the Romanian Ministry for Education and Research.